\documentclass[graybox]{svmult}

\usepackage[justification=centering]{caption}
\usepackage{subcaption}
\usepackage{type1cm}        
\usepackage{makeidx}         
\usepackage{graphicx}        
\usepackage{multicol}        
\usepackage[bottom]{footmisc}
\usepackage{array}
\usepackage{newtxtext}       %
\usepackage{newtxmath}       

\usepackage{todonotes} 

\makeindex             

\begin{document}

\title*{Can Ensemble of Classifiers Provide Better Recognition Results in Packaging Activity?}
\author{A.H.M. Nazmus Sakib, Promit Basak, Syed Doha Uddin, Shahamat Mustavi Tasin, and Md Atiqur Rahman Ahad}
\authorrunning{Sakib et el.}
\institute{A.H.M. Nazmus Sakib \at Electrical and Electronic Engineering, University of Dhaka, Bangladesh \newline \email{nazmussakib2970@gmail.com}
\and Promit Basak \at Electrical and Electronic Engineering, University of Dhaka, Bangladesh
\newline \email{basakpromit@gmail.com}
\and Syed Doha Uddin \at Electrical and Electronic Engineering, University of Dhaka, Bangladesh \newline \email{doha.yeamin@gmail.com}
\and Shahamat Mustavi Tasin \at Electrical and Electronic Engineering, University of Dhaka, Bangladesh \newline \email{tasin.mustavi@gmail.com}
\and Md Atiqur Rahman Ahad \at Electrical and Electronic Engineering, University of Dhaka, Bangladesh \newline Department of Intelligent Media, Osaka University, Japan \newline \email{atiqahad@du.ac.bd}
}

\maketitle

\abstract{Skeleton-based Motion Capture (MoCap) systems have been widely used in the game and film industry for mimicking complex human actions for a long time. MoCap data has also proved its effectiveness in human activity recognition tasks. However, it is a quite challenging task for smaller datasets. The lack of such data for industrial activities further adds to the difficulties. In this work, we have proposed an ensemble-based machine learning methodology that is targeted to work better on MoCap datasets. The experiments have been performed on the MoCap data given in the Bento  Packaging Activity Recognition Challenge 2021. \textit{Bento} is a Japanese word that resembles \textit{lunch-box}. Upon processing the raw MoCap data at first, we have achieved an astonishing accuracy of 98\% on 10-fold Cross-Validation and 82\% on Leave-One-Out-Cross-Validation by using the proposed ensemble model.}

\section{Introduction}
\label{sec:1}
Human Activity Recognition has been one of the major concentrations for researchers for over a decade. Previously, human activity recognition tasks only used data from geospatial sensors such as accelerometers, gyroscopes, GPS sensors, etc. \cite{Lara2013}. But in the last few years, Skeleton-based Human Action Recognition (SHAR) became quite popular because of its better performance and accuracy \cite{Juan2018, Zhu2016, Cippitelli2016, Sarker2021}. In SHAR, the human skeleton is typically represented by a set of body markers which are tracked by several specialized cameras. In such work, computer vision or sensor data can also be used. In the case of sensor data, the use of motion capture, kinematic sensors, etc. are prominent \cite{Ahad2021}. Although Skeleton-based data is already being used widely in many cases, such applications are quite rare in fields such as packaging, cooking \cite{cook2020}, nurse care \cite{Basak2020}, etc. Among these fields, packaging activity recognition can be very effective in the industrial arena and can solve multi-modal problems like industrial automation, quality assessment, and reducing errors. 

Packaging activity recognition is a relatively newer field of SHAR. Hence, scarcity of data and lack of previous examples are some of the main problems of this task. Usually, in such tasks, items are put on a conveyor belt and packaged by a subject. The items to be put on the conveyor belt are dictated by the company. Hence, the location of the item on the belt is also important as it produces new scenarios for different positions. The size of the dataset to perform such experiments is another important factor because small datasets often limit the scope of work. In this paper, our team \textit{Nirban} proposes an approach to solve these issues and detect such activities in the Bento Packaging Activity Recognition Challenge 2021 \cite{bento2021, Adachi2021}. 

The dataset used in this work is based on Bento box packaging activities. Bento is a single-serving lunch box originated in Japan. The dataset contains motion capture data with 13 body markers with no previous preprocessing. The raw motion capture data is first preprocessed, and then a total of 2400 features are extracted. Furthermore, feature selection is used to select the best 396 features based on “mean decrease in impurity” and chi-square score. Then, the processed data are trained on several classical machine learning models and their performances are evaluated using 10-fold Cross-Validation (CV) and Leave-One-Out-Cross-Validation (LOOCV). Lastly, an ensemble of the best five models is done to generate predictions on the test data. Deep Learning (DL) methods were also considered, in which case raw data was fed to 1 dimensional CNN, LSTM, and Bidirectional LSTM. The results of all the approaches and models are included in this paper.

The rest of the paper is organized as follows: Section \ref{sec:2} describes the previous works that are relevant to the approach described here. Section \ref{sec:3} provides a detailed description of the dataset, including its settings and challenges. Section \ref{sec:4} entails the detailed methodology used in this work. Section \ref{sec:5} describes the results and analysis of the results, including the approach, as well as the future scopes of this work. Finally, the conclusion is drawn in section \ref{sec:7}.

\section{Related Work}
\label{sec:2}

Several pieces of research regarding SHAR have been carried out where the dataset had motion capture data. Picard et al. \cite{Picard2021} used motion capture data to recognize cooking activities. The method achieved a score of 95\% on Cross-Validation of the training set. In this work, a subject has been visualized as a stickman using the MoCap data. For temporal information to be taken into consideration, an HMM model was used in post-processing to get a better result. It should be noted that the dataset had few wrong labels, which were manually labeled here and data was shuffled which was also ordered before training. It helped them reach such a high accuracy. For different classes, two specialized classifiers were used and their results were merged.

Image-based approaches have been also observed in the industrial packing process of range hood (meaning kitchen chimney) \cite{Chen2020}. In this case, Local Image Directed Graph Neural Network (LI-DGNN) is used on a set of different types of data. The dataset includes RGB videos, 3D skeleton data extracted by pose estimation algorithm AlphaPose, local images, and bounding-box data of accessories. However, since it uses local images from video frames, it is subjected to object occlusion and viewpoint changes if used solely. Also, as the items are needed to be tracked continuously, it causes a bottleneck in using local images. As a result, a combination of local images and other sensor data is required.

An important observation regarding SHAR works is the size of the datasets. In most cases, the dataset is large enough to experiment with deep learning approaches. Deep learning methods are expected to outperform classical machine learning models with hand-crafted features \cite{Hossain2021}. In this case, a large dataset is advantageous for a data-driven method. However, such an approach is not expected to work well on smaller datasets as given in the Bento Packaging Activity Recognition Challenge 2021.

\section{Dataset}
\label{sec:3}

\subsection{Data Collection Setup}
\label{subsec:1}
\begin{figure}[t]
\centering
\captionsetup{justification=centering}
      \includegraphics[scale=.25]{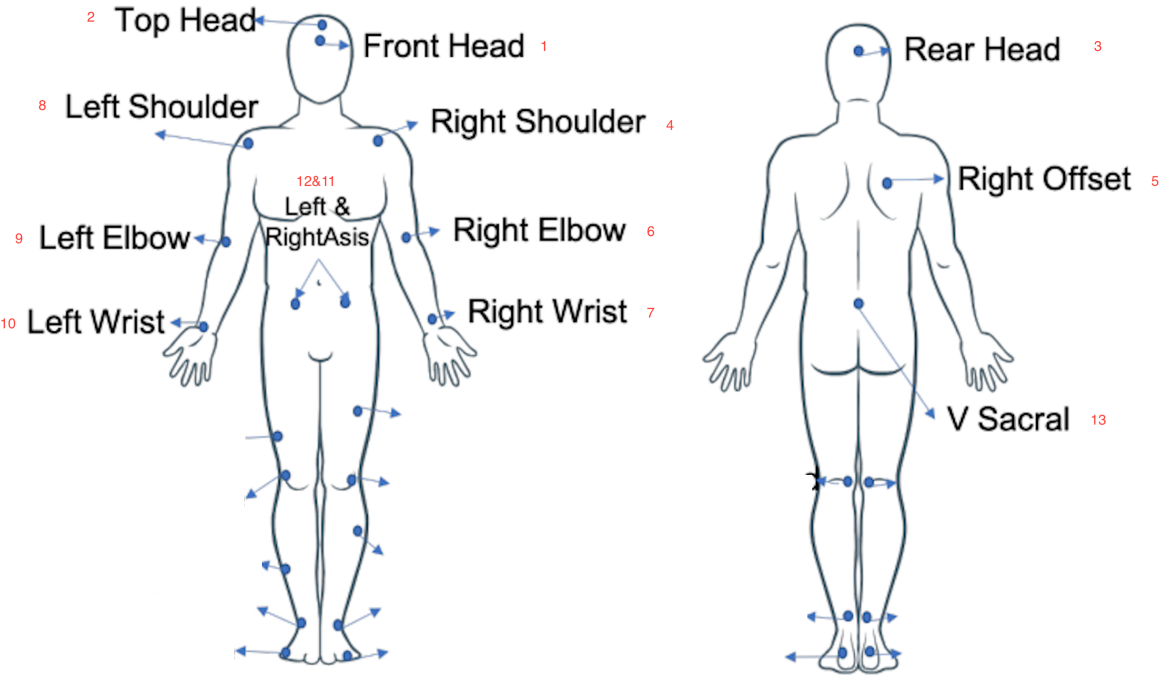}
    \caption{Position of the markers (Source: \url{https://abc-research.github.io/bento2021/data/})}
    \label{figure:1} 
\end {figure}

In any activity recognition challenge, the environment in which data is collected is very crucial. For the Bento Packaging Activity Recognition Challenge 2021, data is collected in the Smart Life Care Unit of the Kyushu Institute of Technology in Japan. Motion analysis company \cite{motionanalysis} has provided the necessary instruments to collect motion capture data. The setup consists of 29 different body markers, 20 infrared cameras to track those markers, and a conveyor belt, where the Bento boxes are passed. Though there were 29 different body markers initially, data for only 13 markers of the upper body are provided for this challenge. The marker positions are shown in Figure \ref{figure:1}.

\begin{figure}[t]
    \centering
      \includegraphics[scale=.50]{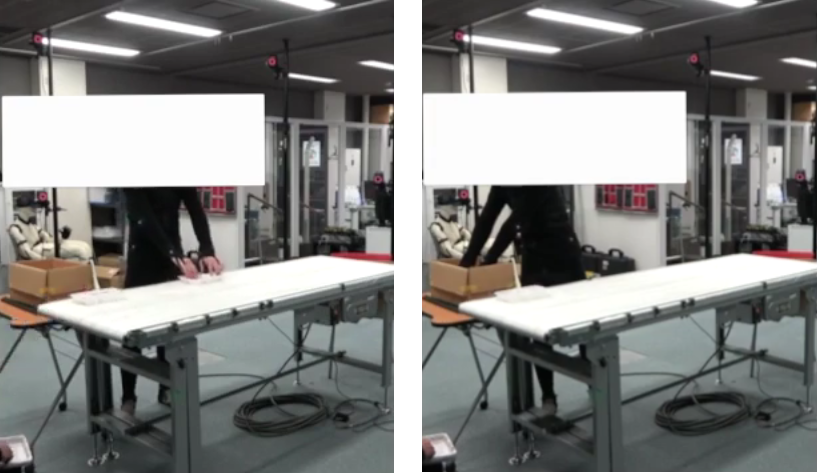}
      
    \caption{Data collection setup. A subject is putting items in a Bento box, which is on the white conveyer belt in the middle. At left-mid, the rectangular box is used to cover the subject's face to retain privacy. (Source: \url{https://youtu.be/mQgCaCjC7fI})}
    \label{figure:2} 
\end {figure}

The data is collected from 4 subjects aged from 20 to 40. While collecting data, empty Bento boxes are passed to the subject using a conveyor belt and the subject has to put three kinds of foods in the box. The data collection setup is given in Figure \ref{figure:2}, where a subject is taking food and putting it in the Bento box. The face is covered with a white rectangular mask to protect privacy. 

\subsection{Dataset Description}
\label{subsec:2}

The dataset for the Bento Packaging Activity Recognition Challenge 2021 consists of activities from five different scenarios necessary for packaging a Bento box. For each scenario, the activity is done in two different patterns which are inward and outward. The name and label for each activity are listed in Table \ref{table:1}.

\begin{table}[ht]
\caption{Activity names and corresponding labels}
\label{table:1}
    \centering
    \begin{tabular}{cc}
    \hline\noalign{\smallskip}
    \textbf{Activity Name}                & \textbf{Label} \\
    \noalign{\smallskip}\svhline\noalign{\smallskip}
    Normal (inward)                       & 1              \\
    Normal (outward)                       & 2              \\
    Forgot to put ingredients (inward)    & 3              \\
    Forgot to put ingredients (outward)   & 4              \\
    Failed to put ingredients (inward)     & 5              \\
    Failed to put ingredients (outward)     & 6          \\
    Turn over Bento box (inward)          & 7              \\
    Turn over Bento box (outward)         & 8              \\
    Fix/rearranging ingredients (inward)  & 9              \\
    Fix/rearranging ingredients (outward) & 10             \\
    \noalign{\smallskip}\hline\noalign{\smallskip}
    \end{tabular}
\end{table}

\begin{figure}[ht]
\centering
\includegraphics[scale=.35]{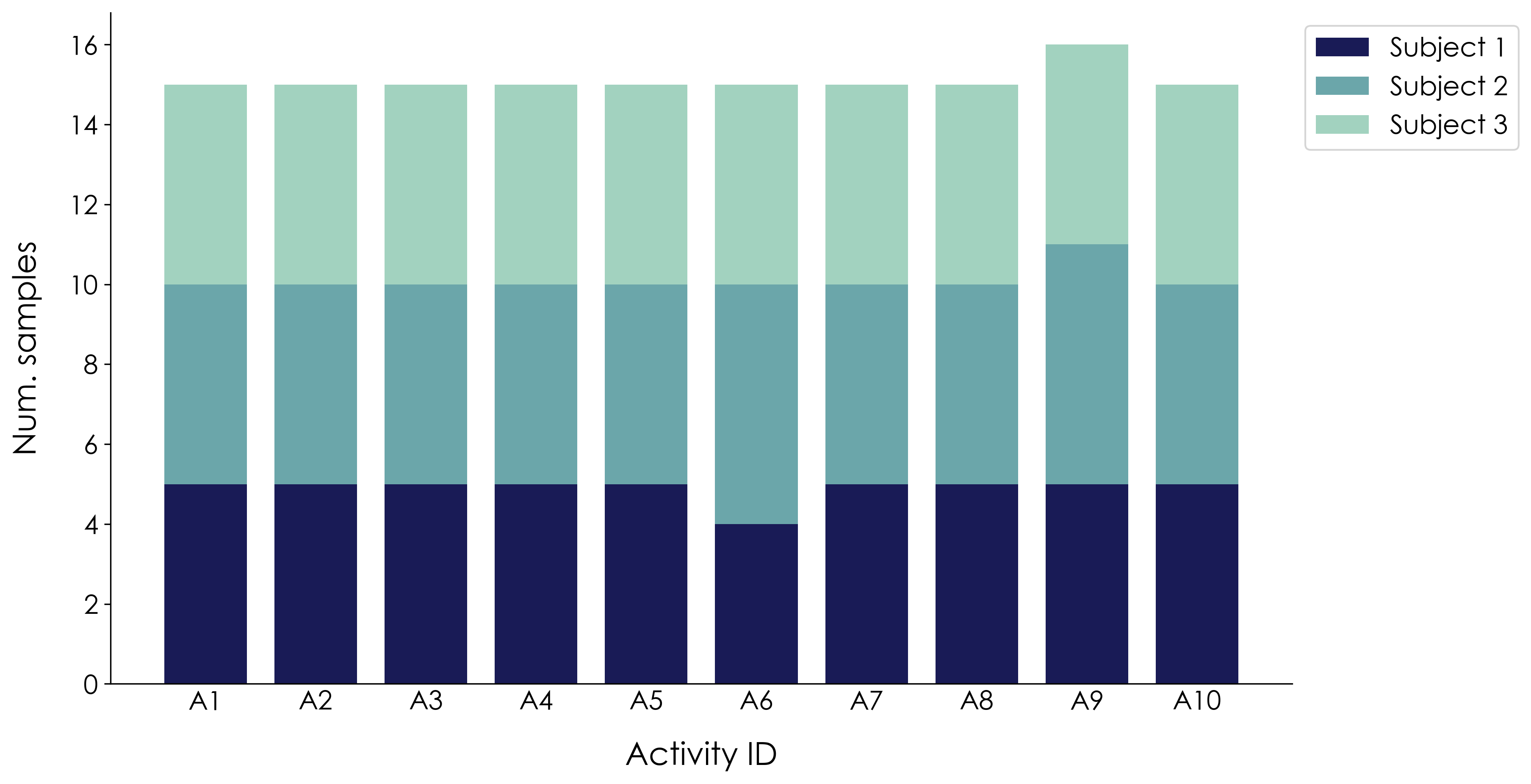}
    \caption{Distribution of samples for every class (A1\textasciitilde A10). Data of 3 subjects (as per the training set) are shown.}
    \label{figure:3} 
\end {figure}

The provided data contains three-dimensional coordinates for each of the body markers sampled at a frequency of 100 Hz. Each subject has performed each activity approximately five times and the duration of each activity is between 50 seconds to 70 seconds. There are a total of 151 training files and 50 test files where each file represents a single activity. From the subject-wise data distributions shown in Figure \ref{figure:3}, it is evident that the dataset is well balanced and each of the subjects has done almost an equal number of different activities. 

\subsection{Dataset Challenges}
\label{subsec:3}

\begin{figure}[ht]
\centering
\includegraphics[scale=.12]{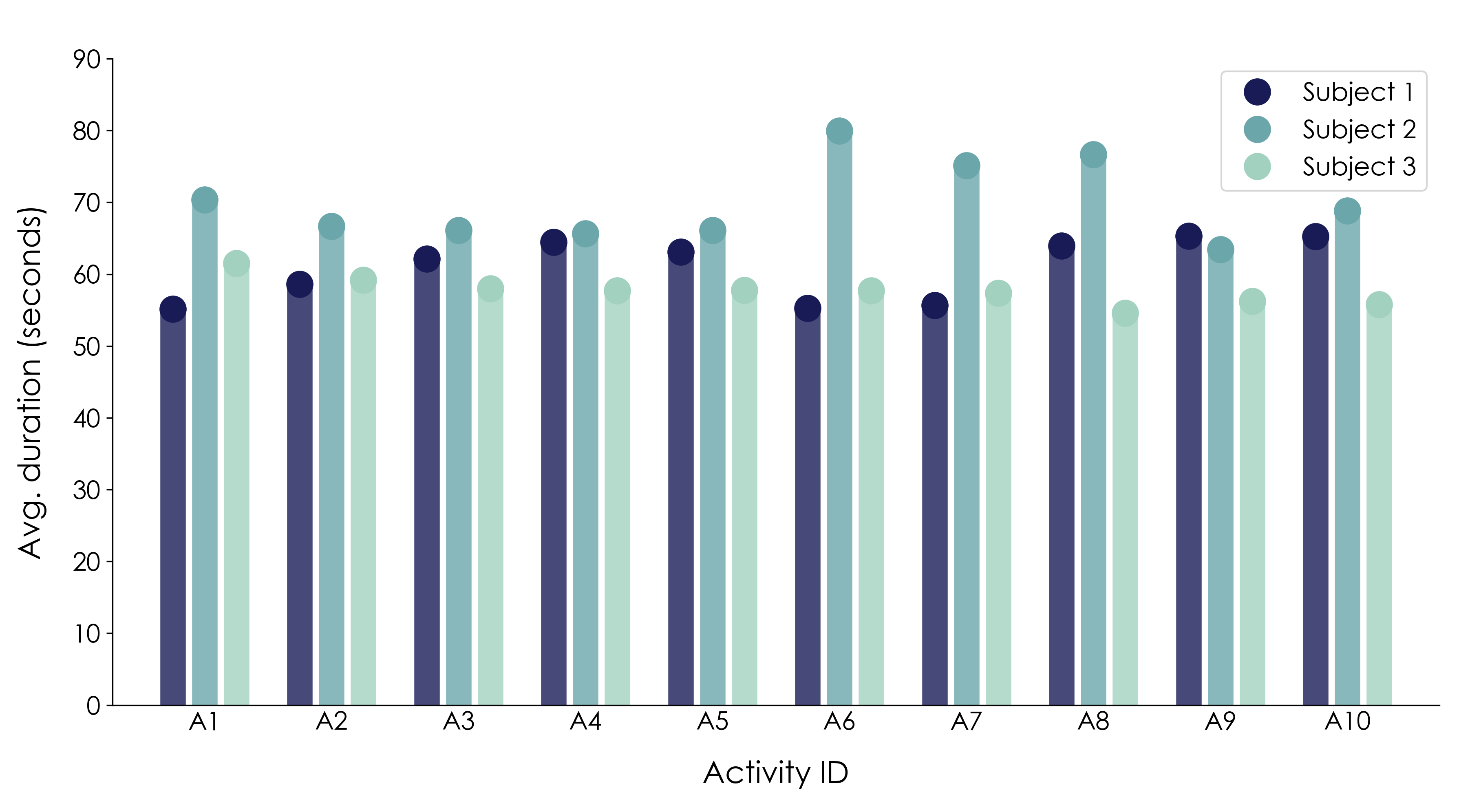}
    \caption{Subject-wise average activity distribution for 3 subjects. Distribution for test subject is unknown.}
    \label{figure:4} 
\end {figure}

In most cases, a real-life data collection setup has some inevitable inconsistencies, which make it challenging to work on. The dataset given in this challenge is not free from this issue too. The biggest challenge of this dataset is the small amount of data. There is only a total of 151 instances in the training set which is extremely low considering the complexity of each activity. This problem makes it very hard to use deep learning models that require large datasets to perform well \cite{Suto2016}. Another difficulty of the dataset is shown in Figure \ref{figure:4}, which compares the average execution time of each activity for different subjects. It is obvious that different subjects have done the activities differently. Subject 2 has taken a significantly longer period to execute the activities in comparison with subjects 1 and 3. This problem is more evident for activities 7, 8, and 9. As the test set contains actions from a different subject who is not present in the training set, this cross-subject inconsistency is likely to take a toll on the overall performance of the model.

However, there are some other problems in the dataset. The data is provided in raw *.csv files rather than any conventional motion capture data format such as, *.bvh, *.htr *.c3d, *.asf, etc. \cite{Meredith2001}. The base positions of the body joints are not provided too. As a result, many important features can not be extracted properly from the files. The setting of data collection is very complex which caused incorrect marker labeling, missing data, and unwanted noises which offered further challenges. Of the 29 markers, data from only 13 markers of the upper body were given. Some of the activities are easily separable if lower body marker data is provided. We have addressed almost every issue in our work, which will be covered in the next sections.

\section{Methodology}
\label{sec:4}

\subsection{Preprocessing}
\label{subsec:4}

The most prominent challenge of this dataset is the low number of instances in the training data. To minimize the problem of data scarcity, we have divided each data into multiple overlapped segments of 20 seconds to 40 seconds. Smaller segments increase the number of instances sacrificing the global trend while larger segments decrease the number of instances retaining the global trend. Hence, we have treated the segment size as one of the hyperparameters and have tuned it to the perfect value. A similar approach is taken on the overlapping rate of two consecutive segments. After completing the preprocessing, we were able to increase the number of instances to a range of 300 to 600 for different combinations of segment size and overlapping rate.

Also, there are some missing values in the dataset. We have interpolated them linearly instead of imputation as the dataset has been resampled at a constant frequency. 
We have extracted features for each segment which will be described in the following sub-section.

\subsection{Stream and Feature Extraction}
\label{subsec:5}

\begin{figure}[ht]
\centering
\includegraphics[scale=.13]{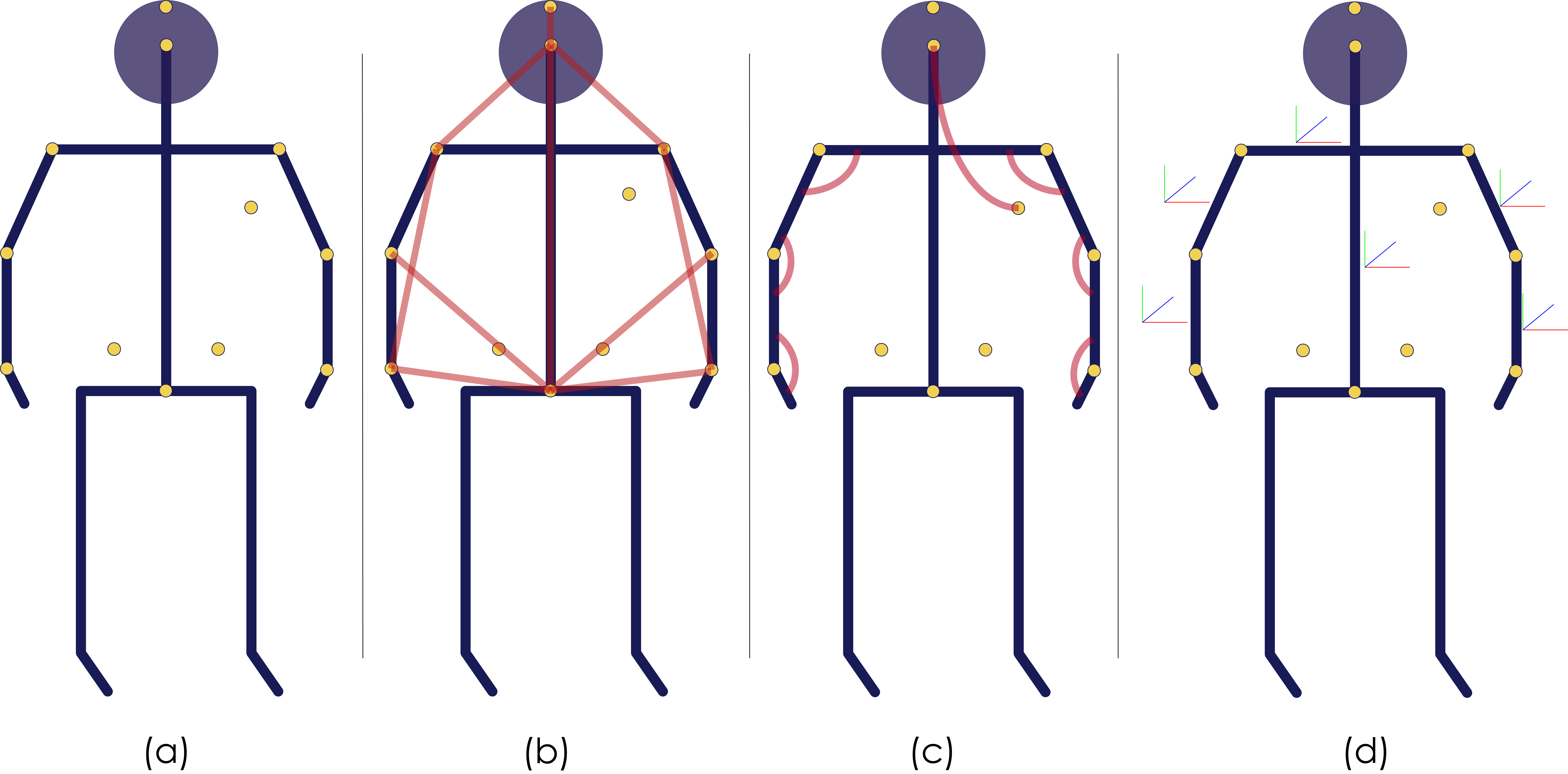}
    \caption{Stream extraction steps: (a) the skeleton model, (b) extraction of distance streams, (c) extraction of joint angle streams, and (d) extraction of planer angle streams}
    \label{figure:5} 
\end {figure}

In this dataset, only the cartesian coordinates of 13 upper body joints are given (Figure \ref{figure:5}). 
So, for each activity, there is a sequence of positions in three axes for all 13 markers. For describing purposes, we will call each of the temporal sequences a stream. This position stream is not enough to describe each activity. So, we have differentiated it repeatedly to get the speed, acceleration, and jerk streams. In Figure \ref{figure:5}, stream extraction steps are mentioned. 

In real life, we move our certain limbs to execute any action and the distances between certain body joints, $d$ are crucial for detecting any action \cite{Cippitelli2016}. From this point of view, we have calculated distances between selected body joints (i.e., the distance between wrist and shoulder, between v-sacral and elbow, between front head and elbows, between the wrists) to create the distance stream. Each distance signifies a separate concept. For example, the distance between the v-sacral and front head helps to determine if the person has bent his/her head or not. The distance is defined as,

\begin{equation}
    d = \sqrt{x^2+y^2+z^2}
\end{equation}

For different activities, the angle between three consecutive joints, $\theta$ and the orientation of the selected bones (i.e., forehand, hand) should be very important. Hence, we have extracted selected joint angle streams and planar angles for bone streams too. In both cases, we have differentiated each stream to get the angular speed streams, according to the following expression,

\begin{equation}
    \theta = \arccos\left ( {\frac{\vec{v_1}}{\left \| v_1 \right \|} \cdot \frac{\vec{v_2}}{\left \| v_2 \right \|}} \right )
\end{equation}

We have synthesized a total of 218 streams after the stream extraction process. The next process is different for RNN-based models and traditional machine learning models. 
RNN-based deep learning models like LSTM can take each stream directly as the input of the network as well-crafted deep learning networks can learn features from data on their own. Hand-crafted feature extraction is unnecessary for these models. But the number of streams is too much for the dataset size. For this reason, we have selected the most important 40 streams for LSTM to train on.

We have observed that the deep learning models performed very poorly because the dataset was so small. So, we have constructed a separate feature extraction pipeline for traditional machine learning models. From each of the selected streams, we have extracted the basic frequency domain features (i.e., median, skew, kurtosis, energy) apart from some statistical features (i.e., median, min, max, standard deviation).

After the execution of the feature extraction pipeline, the feature set became quite large compared to the dataset. So, we had to remove a considerable portion of the feature set to prevent overfitting. We have used the mean decrease in impurity \cite{Nguyen2015} and chi-square techniques \cite{Suto2016} to select the most significant 496 features for the later workflow.

\subsection{Model Selection and Post-processing}
\label {subsec:6}
\begin{figure}[ht]
\centering
\includegraphics[scale=.12]{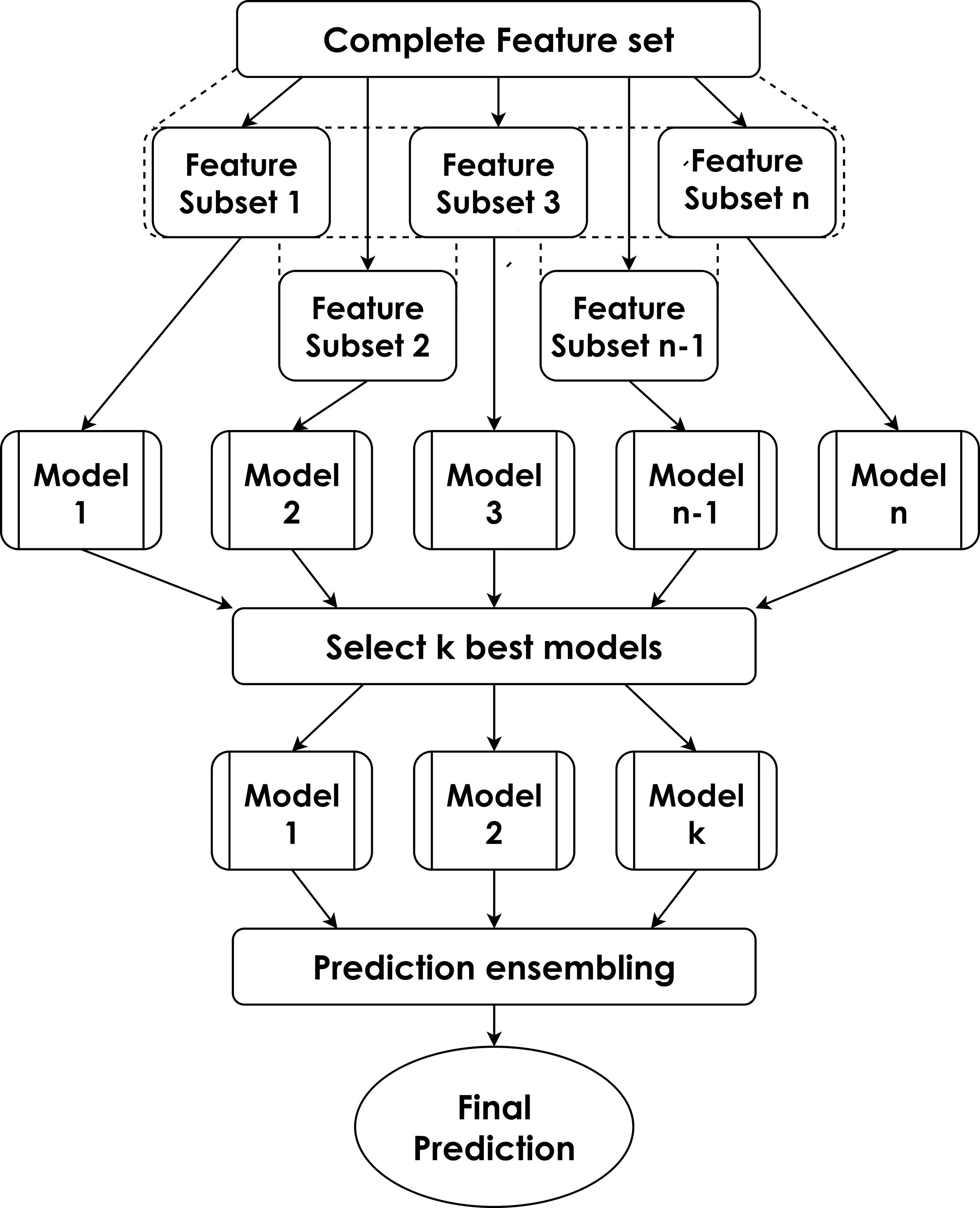}
    \caption{Our propesed ensemble-based framework}
    \label{figure:6} 
\end {figure}

Even after taking a much smaller set of features through the feature selection methods, the number of features remains quite high compared to the dataset size. After preprocessing, the highest number of instances we have produced is less than 600, where the number of features is already as many as 496. This data-to-feature ratio will highly likely lead to overfitting. So, we have proposed a model ensembling system to solve this problem (Figure \ref{figure:6}). 

First, we have divided the feature set into 13 overlapped feature subsets each of which contained approximately 150-250 features. For each model, we have trained on different feature subsets and evaluated its result on both 10-fold CV and LOOCV. We have selected the top five models based on both the evaluation scores and added a majority voting layer on top of each of the models' predictions. 
The intuition behind the proposed ensemble system is that the reduced feature set will make the models less prone to overfitting and majority voting will combine all predictions and give us a final prediction that considers the full feature set \cite{Bayat2014}.

\section{Results and Analysis}
\label{sec:5}
\begin{figure}[ht]
\centering
    \captionsetup{justification=centering}
    \begin{subfigure}{\textwidth}
    \centering
    \includegraphics[scale=.116]{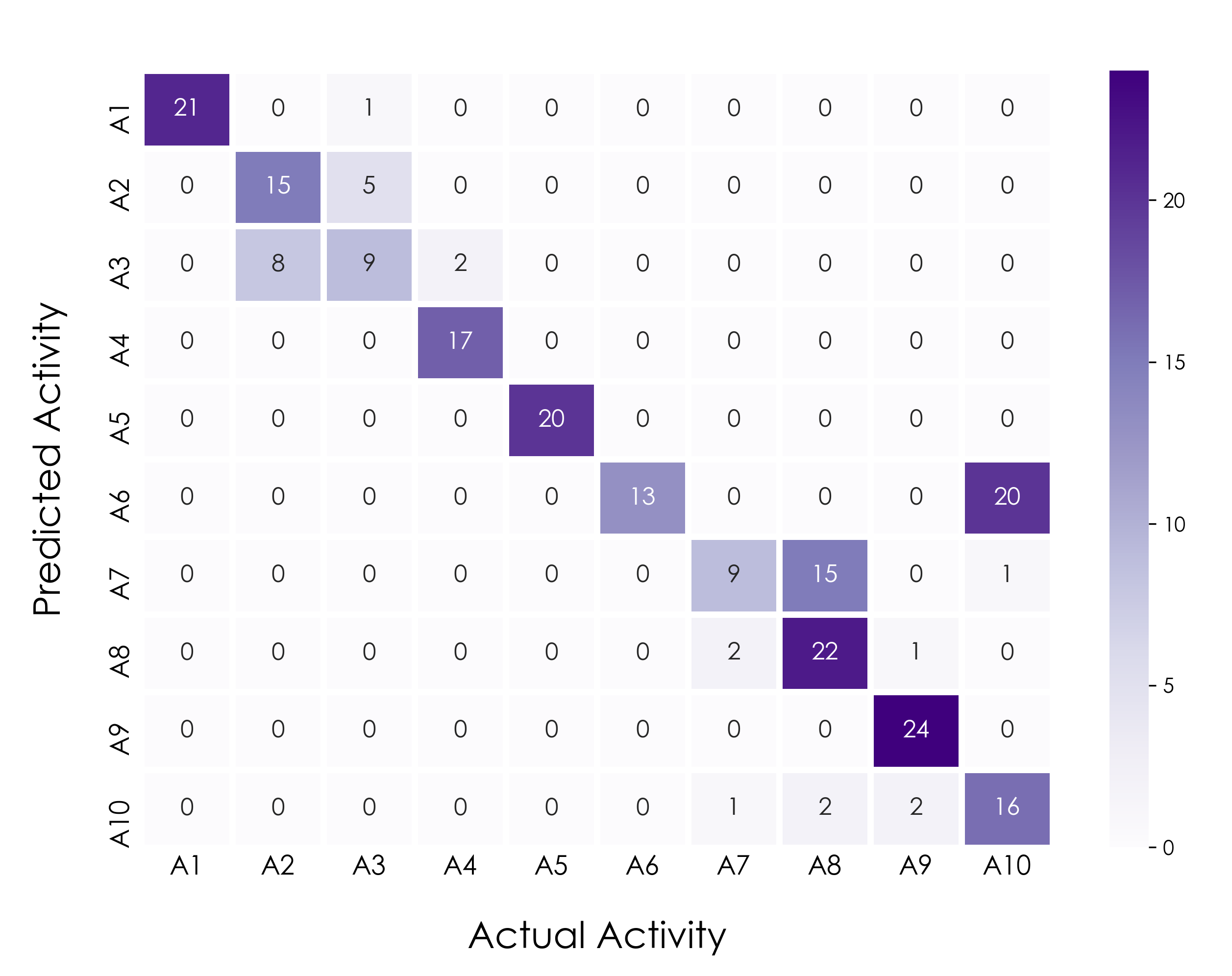}
        \caption{Confusion matrix for the best model after LOOCV.}
        \label{figure:7a} 
    \end{subfigure}

    \begin{subfigure}{\textwidth}
    \centering
    \includegraphics[scale=.116]{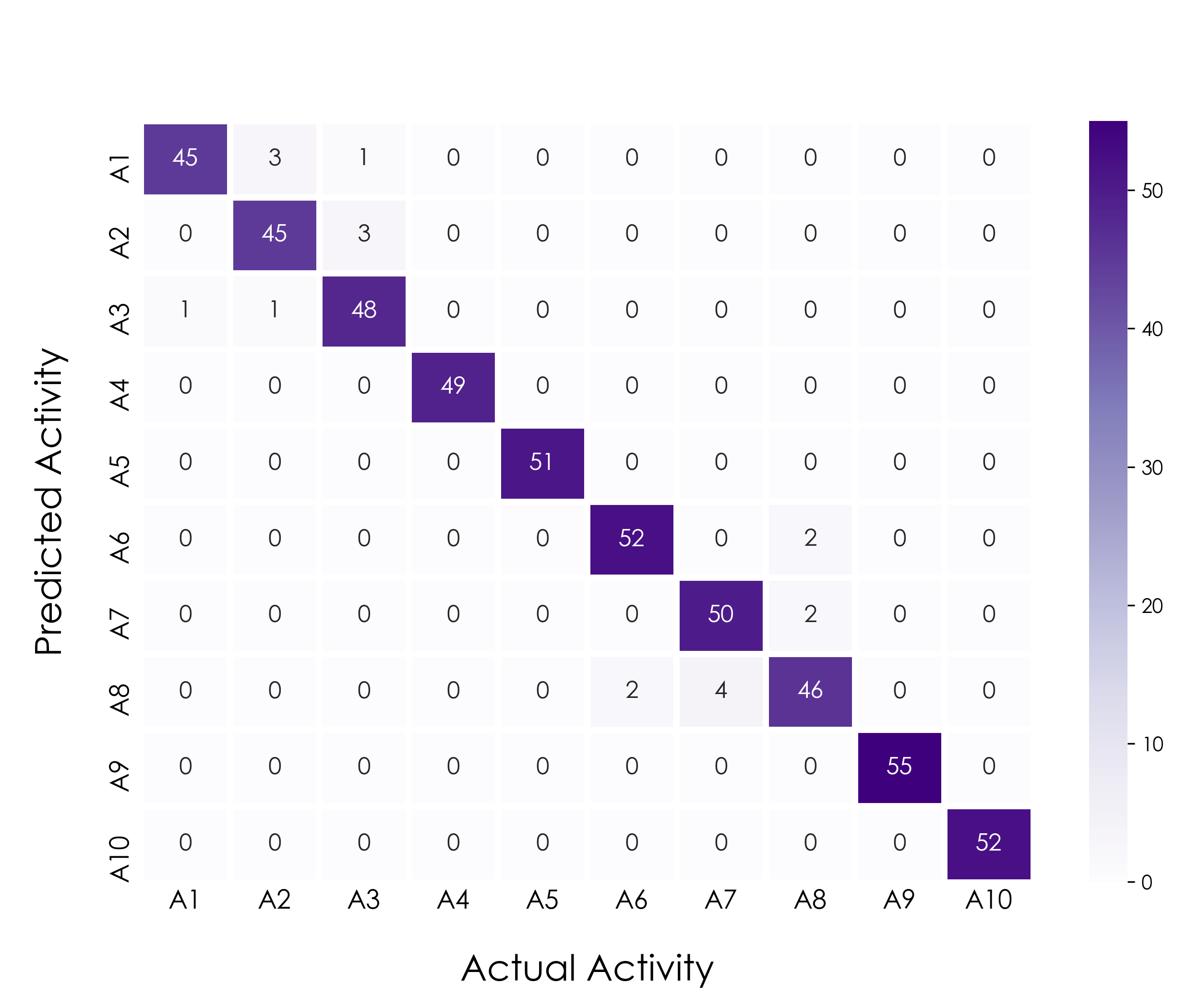}
        \caption{Confusion matrix for the best model after 10-fold CV.}
        \label{figure:7b} 
    \end{subfigure}
\caption{Confusion Matrices}
\label{figure:7}
\end {figure}

We have used different tuned models and evaluated them based on 10-fold CV and LOOCV. We have found the Extra Trees Classifier to perform best on our framework. The detailed results are portrayed in Table \ref{table:2}.
The confusion matrices of the best model found after LOOCV and 10-fold CV are depicted in Figure \ref{figure:7}.

\begin{table}[ht]
\caption{Accuracy of different models}
\label{table:2}
\centering

\begin{tabular}{>{\centering\arraybackslash}m{5cm} >{\centering\arraybackslash}m{3cm} >{\centering\arraybackslash}m{3cm}}
    \noalign{\smallskip}\svhline\noalign{\smallskip}
\textbf{Model} & \textbf{10-fold CV Accuracy} & \textbf{LOOCV Accuracy} \\
    \noalign{\smallskip}\svhline\noalign{\smallskip}
Naive Bayes                     & 0.84      & 0.64  \\
Support Vector Machine (SVM)    & 0.92      & 0.73  \\
Random Forest Classifier (RFC)  & 0.96      & 0.77  \\
Extra Trees Classifier (ETC)    & 0.96      & 0.78  \\
LightGBM                        & 0.94      & 0.74  \\
XGBoost                         & 0.88      & 0.67  \\
Long Short-Term Memory (LSTM)   & 0.61      & 0.37  \\
\textbf{Ensemble of 4 RFC models and 1 ETC model} 
                & \textbf{0.98} & \textbf{0.82}      \\
    \noalign{\smallskip}\svhline\noalign{\smallskip}
\end{tabular}
\end{table}

The highest accuracy we have achieved in this work is 98\% for 10-fold CV and 82\% for LOOCV. Though the result we obtained is not perfect, this is a very competitive result considering the dataset challenges. There are several reasons behind it. As we can see from Table \ref{table:2}, the evaluation on LOOCV is pretty much less than that of the 10-fold CV score. The reason behind it is the lower number of samples for each activity. Each subject has only 5 samples for each activity which is too low for generalizing on another subject and this leads to a significantly lower score on LOOCV.  

Also from Table 2, we can see that some models have done significantly poorer than other models. LSTM has performed the worst because of the small dataset size. Deep learning models generally need a lot of data to obtain a generalized performance on data. In this case, the dataset size is so small that LSTM has performed even lower than the baseline model, which is the Naive Bayes Classifier. Gradient boosting models, XGBoost and LightGBM also suffered a lot from this problem and do not perform very well.

If we look at the confusion matrix for LOOCV in Figure \ref{figure:7a}, we can notice that the model can not differentiate between classes 2 and 3, 7 and 8, 6 and 10. It is because different subjects carried out the activities differently to some extent. This problem is also evident in Figure \ref{figure:4}, which depicts the average activity execution time for different subjects. For example, for activity 6, subject 1 and subject 3 took 60 seconds on average to perform, but subject 2 took more than 80 seconds to perform the same task. This problem is reflected in the confusion matrix as we can see the model was severely confused between activity 6 and activity 10. Also, activities 2 and 3 are typically distinguishable, but there are some confusions observed in the confusion matrix for LOOCV. On the other hand, activities 7 and 8 are similar but confusions are observed in this case too. Video data of these activities might help to solve these issues in the future.
Despite different challenges and complications of the dataset, our proposed procedure has managed to achieve a quite promising result.

\section{Conclusion}
\label{sec:7}
In this paper, we have provided a method to tackle the challenges of activity recognition in an industrial setting. Though human activity recognition is a very popular field and a wide variety of work has been done, our work still manages to provide a solution in a less explored arena of this field.
After comparing various methods used in previous works by applying them to our dataset, we have decided to use a hand-crafted feature-based solution for our final approach. We have calculated various streams such as speed, joint angle, marker distance from the given data. Furthermore, we have used segmentation with overlap to increase the amount of the data. After that, we have extracted statistical features from each stream. The features are then used to train different machine learning models. After tuning the models and evaluating them using LOOCV and 10-fold CV methods, five best-performing models (four Random Forest Classifiers and one Extra Tree Classifier) were selected. The models are used to make predictions on different segments generated from the files of the test dataset, and a majority voting system among the models generates the final predictions.

Our method provides a good amount of precision, but further improvement is still possible. We have experimented with quaternion data generated from provided three-dimensional coordinates, but could not manage to obtain significant improvement through its use. Finding a system to incorporate this data might provide better accuracy. We have also explored different deep learning methods (i.e., Temporal Convolutional Network, LSTM-based encoder-decoder, Bidirectional LSTM-based network), but they perform poorly. We believe the reason for this is the small size of the dataset. Deep learning approaches have the potential to perform better than machine learning approaches, provided that more data is collected. It will also be able to provide end-to-end solutions in contrast to our hand-crafted feature-based solution, which will streamline its integration in the industry. Thus, the collection of more data and exploring deep learning approaches on the data should be strongly considered for future works.

\bibliographystyle{spmpsci}
\bibliography{bibtex}
\pagebreak
\section*{Appendix}
\begin{table}[ht]
\caption{Miscellanious Information}
\label{table:misc}
\centering
\setlength\extrarowheight{2pt}
\begin{tabular}{>{\centering\arraybackslash}m{3.5cm}|>{\centering\arraybackslash}p{5cm}}
    \hline
    \multicolumn{1}{>{\centering\arraybackslash}m{3.5cm}}{\textbf{Information Heading}} & \multicolumn{1}{>{\centering\arraybackslash}m{5cm}}{\textbf{Description}}     \\
    \hline
Used Sensor Modalities           & Motion Capture (MoCap)                                                    \\
    \hline
Features Used                    & As described in section \ref{subsec:5}                                     \\
    \hline
Programming Language             & Python 3.8                                                                     \\
    \hline
    Packages Used                    & Pandas, Numpy, SciPy, Scikit-learn, Xgboost, Tensorflow, Keras             \\ 
    \hline
Machine Specifications           & Intel(R) Core(TM) i5-7200U CPU @ 2.50GHz, 8 GB RAM                           \\
    \hline
Training and Testing Time        & Time to process and train on full training data - 12 minutes \\
                                 & Time to predict on full test data - 3 minutes    \\ 
    \hline
\end{tabular}
\end{table}
\end{document}